\documentclass[letterpaper]{article} 
\usepackage{aaai2026}  
\usepackage{times}  
\usepackage{helvet}  
\usepackage{courier}  
\usepackage[hyphens]{url}  
\usepackage{graphicx} 
\usepackage{natbib}  
\usepackage{caption} 
\usepackage{algorithm}
\usepackage{algorithmic}

\usepackage{newfloat}
\usepackage{listings}
\DeclareCaptionStyle{ruled}{labelfont=normalfont,labelsep=colon,strut=off} 
\floatstyle{ruled}
\newfloat{listing}{tb}{lst}{}
\floatname{listing}{Listing}
\usepackage{booktabs}
\usepackage{multirow}
\usepackage{amsmath}
\usepackage{amssymb}

\newcommand{\midsepremove}{\aboverulesep = 0.3025mm \belowrulesep = 0.492mm}
\newcommand{\midsepdefault}{\aboverulesep = 0.605mm \belowrulesep = 0.984mm}
\newcommand{\argmin}{\mathop{\mathrm{argmin}}}
\newcommand{\argmax}{\mathop{\mathrm{argmax}}}
\newcommand{\norm}[1]{\left\|#1\right\|}

\title{Class-Partitioned VQ-VAE and Latent Flow Matching for Point Cloud Scene Generation}
\author {
    Dasith de Silva Edirimuni, 
    Ajmal Saeed Mian
}
\affiliations {
    The University of Western Australia\\
    35 Stirling Highway, \\
    Perth, WA 6009 Australia \\
    \{dasith.desilva, ajmal.mian\}@uwa.edu.au
}

\begin{document}

\maketitle

\begin{abstract} 
Most 3D scene generation methods are limited to only generating object bounding box parameters while newer diffusion methods also generate class labels and latent features. Using object size or latent feature, they then retrieve objects from a predefined database. For complex scenes of varied, multi-categorical objects, diffusion-based latents cannot be effectively decoded by current autoencoders into the correct point cloud objects which agree with target classes. We introduce a Class-Partitioned Vector Quantized Variational Autoencoder (CPVQ-VAE) that is trained to effectively decode object latent features, by employing a pioneering \textit{class-partitioned codebook} where codevectors are labeled by class. To address the problem of \textit{codebook collapse}, we propose a \textit{class-aware} running average update which reinitializes dead codevectors within each partition. During inference, object features and class labels, both generated by a Latent-space Flow Matching Model (LFMM) designed specifically for scene generation, are consumed by the CPVQ-VAE. The CPVQ-VAE's class-aware inverse look-up then maps generated latents to codebook entries that are decoded to class-specific point cloud shapes. Thereby, we achieve pure point cloud generation without relying on an external objects database for retrieval. Extensive experiments reveal that our method reliably recovers plausible point cloud scenes, with up to 70.4\% and 72.3\% reduction in Chamfer and Point2Mesh errors on complex living room scenes. 
\end{abstract}

\begin{links}
    \link{Code and extended paper}{https://github.com/ddsediri/CPVQ-VAE-LFMM}
\end{links}

\begin{figure}[!t]
    \centering
    \includegraphics[width=\linewidth,trim=4 4 4 4,clip]{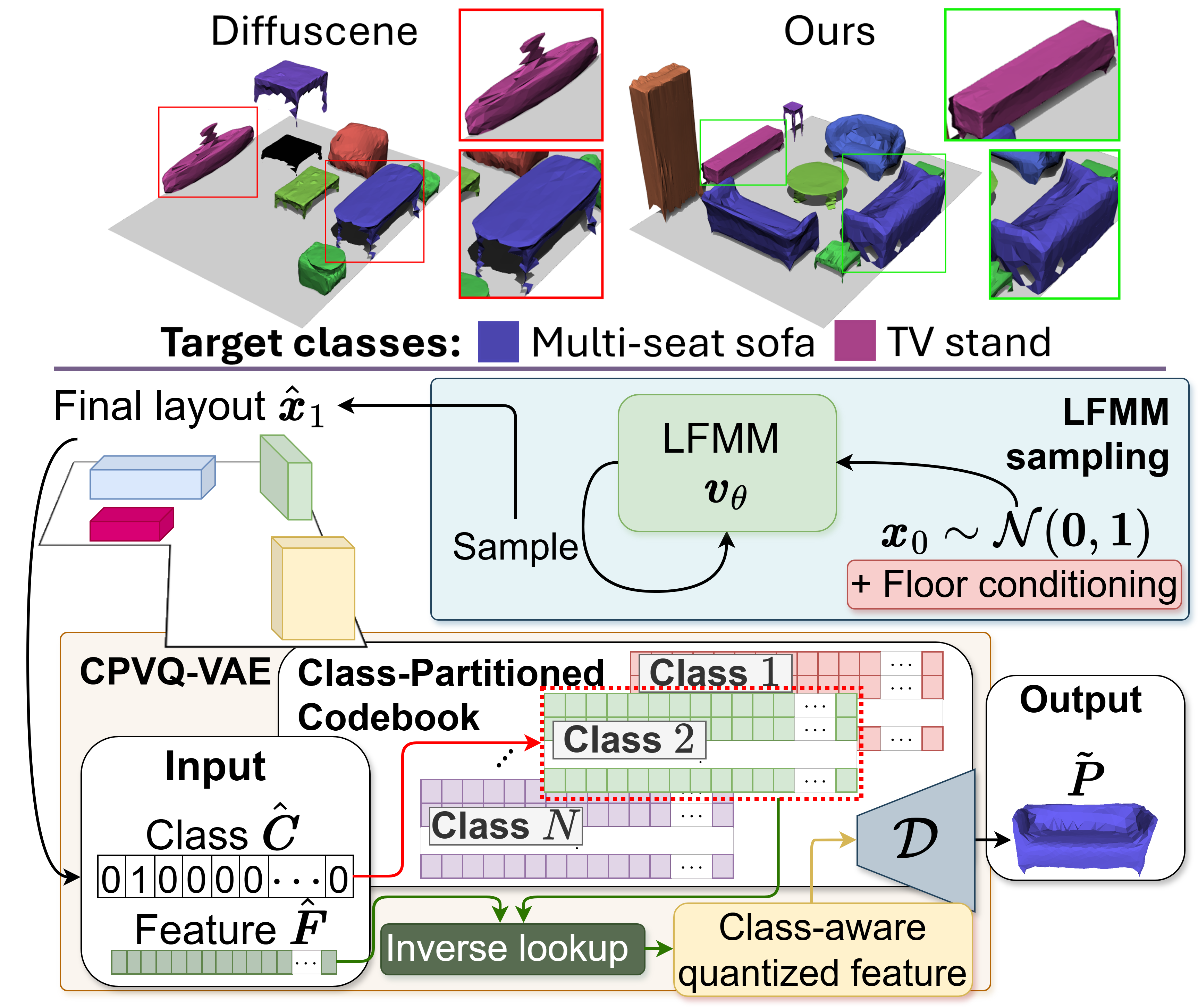}
    \caption{\textbf{Top}: Diffuscene, together with a pretrained VAE, generates object latents resulting in poorly decoded point clouds which \emph{frequently disagree} with the object's target class. Our Latent Flow Matching Model (LFMM) produces more robust latents which are \emph{correctly decoded} by the Class-Partitioned VQ-VAE (CPVQ-VAE), that exploits class labels from the LFMM to look-up the corresponding codevectors. \textbf{Bottom}: Overview of our proposed point cloud scene generation framework. The LFMM generates object bounding box parameters, classes and respective features for the entire scene. Thereafter, the CPVQ-VAE decodes object features, given the generated object class and feature.}
    \label{fig:overview}
\end{figure}

\section{Introduction}

Architectural advancements to neural networks such as the U-Net \cite{Ronnenberger-U-Net} and Transformer \cite{Vaswani-Transformer} have driven rapid progress in text and image generation. This was accelerated by the introduction of ChatGPT \cite{Radford-ChatGPT} and similar variants for natural language processing while latent diffusion models such as StableDiffusion, inspired by \citet{Rombach-Stable-Diffusion}, led the surge in image generation. By contrast, the maturity of 3D generation methods lags behind significantly and they fall under two main categories: 1) object generation and 2) scene generation. Of particular interest are point clouds as they are a compact representation of 3D data that can be efficiently processed compared to other representations such as voxel grids \cite{Qi-PointNet, Qi-PointNet++}. Contemporary object generation methods have mainly employed Continuous Normalizing Flows \cite{Chen-Neural-ODE, Yang-PointFlow} Diffusion \cite{Song-Score-Matching, Luo-Diffusive} and Rectified Flows \cite{Liu-RectifiedFlow, Wu-PointStraightFlow, Li-TripoSG} as the underlying generative mechanisms. Meanwhile, 3D scene generation methods seek to generate multiple objects within a single, plausible, scene. However, current point cloud scene generation methods fall short of producing target point clouds. Instead, they produce the latent shape code of each object point cloud and use this latent to retrieve the corresponding object mesh from a database by minimizing the $L_2$ norm between generated features and database features, the latter must be filtered by class as part of a post-processing step. 

Notably, Diffuscene \cite{Tang-Diffuscene} treats each scene as a stack of object vectors. Each vector is a concatenation of an object's bounding box parameters (centroid translation, rotation, size), class vector and latent code which are generated via a diffusion model. After applying a post-processing step that filters the objects database by generated class, the latent feature is used to retrieve the closest object from the filtered database. This method can be extended to a pure point cloud generation method by directly decoding latents using the pre-trained variational autoencoder (VAE), which is the same used to produce target latents for training. However, we identify one key shortcoming: for complex scene types, such as living and dining rooms, object features generated directly by the diffusion model cannot be effectively decoded by the VAE to produce valid shapes. In fact, Diffuscene generates inexact latents which frequently disagree with the target classes, leading to incorrectly decoded shapes as shown in Fig. \ref{fig:overview}. 

Conversely, methods such as ATISS \cite{Paschalidou-ATISS} and DeBaRa \cite{Maillard-Debara} completely forgo generating object latents and utilize the regressed object sizes to filter the closest object from the database. Based on these observations, we surmise that shape latents generated from a diffusion process must follow the correct class conditioning while also producing a reliable enough latent that can be decoded to a sensible shape. Additionally, Diffuscene takes a comparatively large time to generate the scene layout. In contrast to point cloud methods, signed-distance field based methods require class conditioning and object-object relational information \cite{Zhai-Commonscenes, Zhai-Echoscene, Wu-SEK}. Moreover, the signed distance field estimation is performed on voxel grids  which is less efficient \cite{Ju-DiffInDScene}. To address these limitations, we propose:
\begin{enumerate}
    \item A \emph{pure} point cloud scene generation method that simultaneously generates object bounding box parameters, class labels and latent features.
    
    \item A novel \emph{Class-Partitioned} VQ-VAE (CPVQ-VAE) with a labeled codebook where each vector belongs to a specific class. During inference, the CPVQ-VAE relies on labels, generated by latent space flow matching, when decoding codevectors into class-consistent point clouds. 
    
    \item A class-aware running average update which minimizes the number of \emph{dead} codebook entries and addresses the problem of codebook collapse within the CPVQ-VAE.
    
    \item A 3D latent space flow matching model (LFMM) that takes significantly fewer sampling steps to converge to plausible scene layouts, while generating object features which can successfully be decoded by the CPVQ-VAE. 
\end{enumerate}

Experiments show our method's superiority in generating complex point cloud scenes, \emph{with up to 70.4\% and 72.3\% reduction} in Chamfer and Point2Mesh distance errors, respectively, on the task of living room scene generation.

\section{Related Works}
\paragraph{Latent representation learning using VQ-VAEs:} \citet{Hinton-Autoencoder} were among the first to introduce autoencoders which were later exploited in tasks such as dimensionality reduction \cite{Hinton-Dimensionality-Reduction}. Variational autoencoders were then proposed \cite{Kingma-VAE, Kingma-Generalized-Scalable-VAE} that took advantage of variational inference and facilitated generative methods. Learning discrete representations of latent features using vector quantization was introduced by \citet{Van-Den-Oord-VQ-VAE} and extended to consider hierarchical codebook quantization \cite{Razavi-VQVAE2} with conditioned PixelCNN priors \cite{van-den-Oord-PixelCNN}. Furthermore, a categorical decoder was proposed for VQ-VAE architectures by \citet{Chorowski-Wavenet}. While a powerful representation learning technique, naive vector quantization methods suffer from the problem of \emph{codebook collapse} \cite{Takida-SQ-VAE} which was addressed using techniques such as residual quantization \cite{Lee-Residual-Quantization}, hierarchical quantization \cite{Takida-HQ-VAE} and online clustering \cite{Zheng-Online-Clustered-Codebook}. Specifically, the online clustering method proposes a simple framework for reinitializing dead codevectors via a running average update. 

\paragraph{Generative processes:}
Score matching of \citet{Hyvarinen-Score}, based on physical diffusion, underpins the revolution in image generation with foundational work being done by \citet{Song-Score-Matching}. Diffusion models were further refined by \citet{Ho-DDPM} and \citet{Song-DDIM} who introduced the denoising diffusion probabilistic and implicit models (DDPM and DDIM), respectively, while \citet{Rombach-Stable-Diffusion} proposed to denoise images in latent space. Building on the work of \citet{Maoutsa-Deterministic-LE}, \citet{Song-SDE-PF-ODE} obtained a deterministic process, named the \emph{probability-flow} (PF-ODE) equation which facilitates the distillation of diffusion models into consistency models \cite{Song-Consistency} yielding straightened paths. Similar to consistency models, rectified flows were proposed by \citet{Liu-RectifiedFlow} to regress optimal transport paths from noisy to clean samples by inferring a constant velocity field at each linearly interpolated state between samples. The stochastic interpolants of \citet{Albergo-Stochastic-Interpolants} and flow matching method of \citet{Lipman-Flow-Matching} are two related methods that subsume rectified flows where the velocity field at interpolated points between samples need not be constant. 

\paragraph{Point cloud scene generation:}

Point cloud shape generation using score matching methods was first proposed by \citet{Cai-ShapeGF} and further extended to use to DDPM sampling along with better shape conditioning by the work of \citet{Luo-Diffusive}. However, these methods focus on single object generation and do not handle multi-categorical data. Score matching techniques have also been adopted for point cloud denoising \cite{Luo-Score-Based-Denoising, Edirimuni-IterativePFN}. Similarly, optimal transport based flow matching has been adopted for point cloud shape generation \cite{Wu-PointStraightFlow} and denoising \cite{Edirimuni-StraightPCF}. Meanwhile, early scene generation methods focused on autoregressive retrieval \cite{Ritchie-ISS-DCGM, Wang-Sceneformer, Paschalidou-ATISS}. \citet{Paschalidou-ATISS}, inspired by previous works \cite{van-den-Oord-Wavenet, Salimans-PixelCNN++}, modeled object attributes with a mixture of logistic distributions. Thereafter, LEGO-Net denoised noisy arrangements of given objects to generate rearranged scenes \cite{Wei-LEGO-Net}, while \citet{Tang-Diffuscene} introduced Diffuscene to generate all object parameters (including class labels and features) via diffusion. Diffuscene specifically uses these latents to improve the object retrieval process. Similarly, building on diffusion based methods, \citet{Maillard-Debara} proposed an EDM sampling driven method to retrieve scene layouts, which requires class conditioning to aid bounding box generation. Several other methods \cite{Zhai-Commonscenes, Zhai-Echoscene, Wu-SEK} reverse diffuse Signed Distance Field representations of 3D data but require class labels, and object-to-object relations (in the form of scene graphs) as prior conditioning. Furthermore, \citet{Ju-DiffInDScene} proposed to reverse diffuse voxel occupancy to generate scenes. Finally, \citet{Feng-CASAGPT} proposed CASAGPT which generates cuboid representations of 3D objects which are then used for retrieval.

\section{Methodology}
In this section, we present a Class-Partitioned Vector Quantized Autoencoder (CPVQ-VAE) and Latent space Flow Matching Model (LFMM) for point cloud scene generation. Our goal is to generate a plausible scene $\hat{\pmb{X}} = \left\{\hat{\pmb{x}}^m~|~ 1 \leq m \leq M\right\}$, which consists of a maximum $M$ objects, positioned and oriented consistently to their target classes. Denoised object bounding box attributes are characterized by a center coordinate translation $\hat{\pmb{T}}^m\in\mathbb{R}^{3}$, a rotation $\hat{\pmb{R}}^m=(\cos(\gamma), \sin(\gamma))\in\mathbb{R}^{2}$, w.r.t. the scene's vertical axis where $\gamma$ is the relative angle, and the object size $\hat{\pmb{S}}^m\in\mathbb{R}^{3}$. Moreover, the class attribute vector $\hat{\pmb{C}}^m\in\mathbb{R}^{N_c}$, with $N_c$ the number of classes for a given scene type, is used to obtain the object class $\hat{c}^m=\argmax_{n\in [N_c]}\hat{\pmb{C}}^m_n$. Finally, latent features are represented by $\hat{\pmb{F}}^m\in\mathbb{R}^{32}$. All attributes form a tuple that describes each object within the scene, $\hat{\pmb{x}}^m =(\hat{\pmb{T}}^m, \hat{\pmb{R}}^m, \hat{\pmb{S}}^m, \hat{\pmb{C}}^m, \hat{\pmb{F}}^m)$.

The object feature $\hat{\pmb{F}}^m$ is subsequently decoded into a point cloud $\tilde{\pmb{P}}^m$ using the CPVQ-VAE decoder $\mathcal{D}$ and $\mathbf{L}(\hat{\pmb{F}}^m, \mathcal{C}, \hat{c}^m)$, an inverse look-up function that we introduce to map $\hat{\pmb{F}}^m$ to codebook vectors within the CPVQ-VAE codebook $\mathcal{C}$. The CPVQ-VAE architecture ensures that generated object point clouds are \emph{class-consistent}, i.e., latent features are decoded to point clouds of the target class $\hat{c}^m$.

\begin{figure}[!t]
    \centering
    \includegraphics[width=\linewidth]{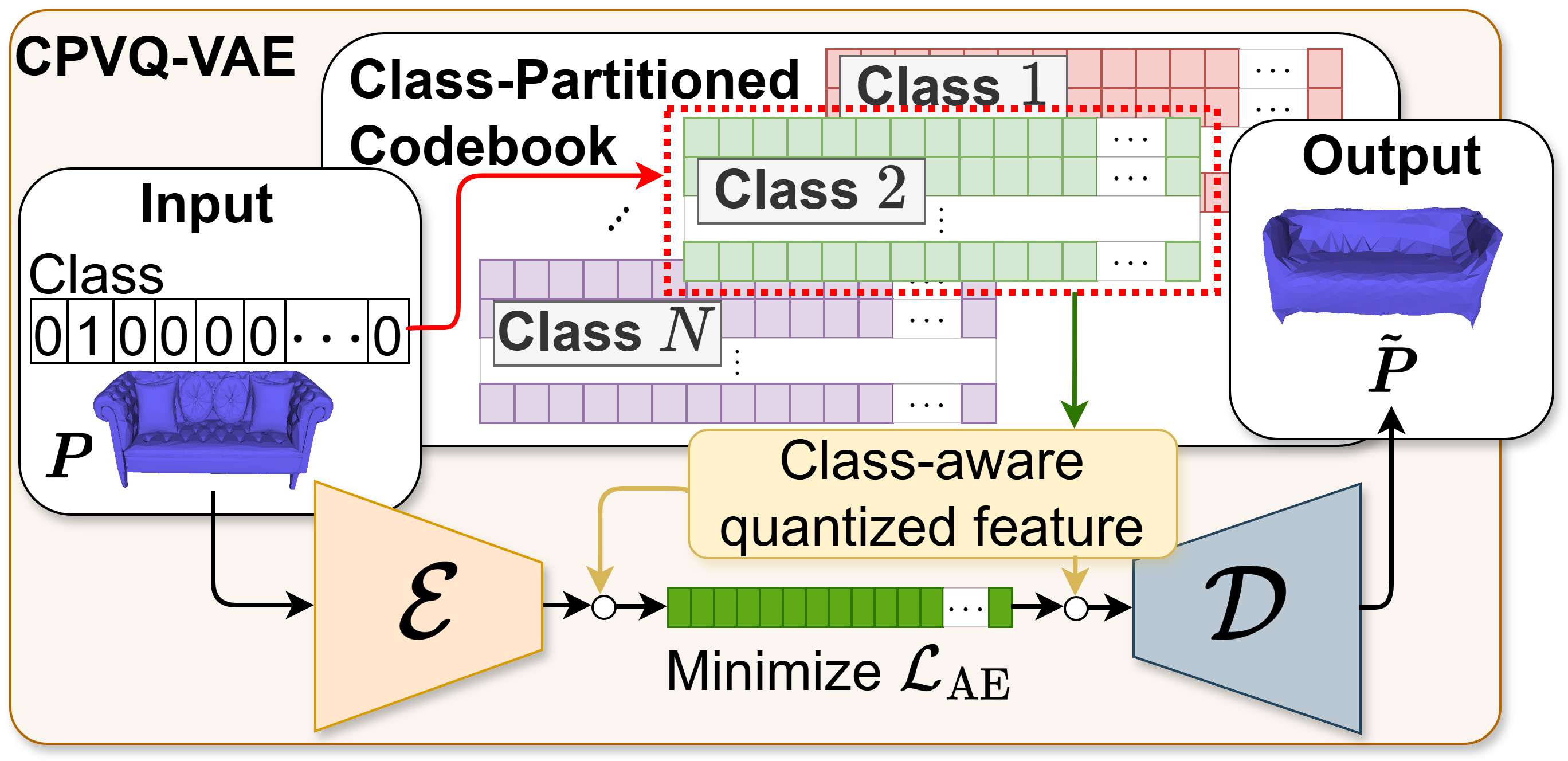}
    \caption{The Class-Partitioned VQ-VAE uses class label inputs to partition the codebook, allowing it to learn class-specific point cloud representations.}
    \label{fig:cpvq-vae-model}
\end{figure}

\subsection{Class-Partitioned VQ-VAE (CPVQ-VAE)}
\label{sec:cpvq-vae}
Previous methods, such as Diffuscene, generate unreliable object latents that frequently produce incorrectly decoded point clouds that do not agree with the target class, or other object properties such as size. We propose a Class-Partitioned VQ-VAE (CPVQ-VAE) to ensure that object latents are mapped to labeled codebook entries in a consistent manner. The key idea is that object classes can be generated by a flow matching process, along with other object parameters such as the object feature. Thereafter, given the CPVQ-VAE's partitioned codebook, the generated class and object feature can be decoded to produce the final 3D point cloud. The proposed CPVQ-VAE architecture is shown in Fig.~\ref{fig:cpvq-vae-model}.

A traditional VQ-VAE consists of an encoder $\mathcal{E}$ and decoder $\mathcal{D}$ along with a codebook $\mathcal{C}=\{(k, \pmb{e}^k)\} \in \mathbb{R}^{N_K \times D_K}$ where $k$ is a given code. Moreover, $\pmb{e}^k$ is the $k$-th codevector, $D_K=128$ is the codevector dimension and $N_K$ is the number of total codevectors. Given a point cloud $\pmb{P}\in\mathbb{R}^{N_P\times 3}$ of $N_P$ points, a VQ-VAE first encodes the point cloud into latent representation $\mathcal{E}(\pmb{P})\in \mathbb{R}^{D_K}$. The quantization process $\mathcal{Q}(\mathcal{E}(\pmb{P}); \mathcal{C})$ maps this encoding to the nearest codevector in $\mathcal{C}$ such that,
\begin{align}
    \mathcal{Q}(\mathcal{E}(\pmb{P}); \mathcal{C})& = k^* =  \argmin_{k\in[N_K]}\norm{\mathcal{E}(\pmb{P}) - \pmb{e}^k}^2_2.
    \label{eq:quantization-process}
\end{align}
The quantized counterpart of $\mathcal{E}(\pmb{P})$ is then $\pmb{z}^q = e^{k^*}$.

\begin{figure*}[!t]
    \centering
    \includegraphics[width=\linewidth]{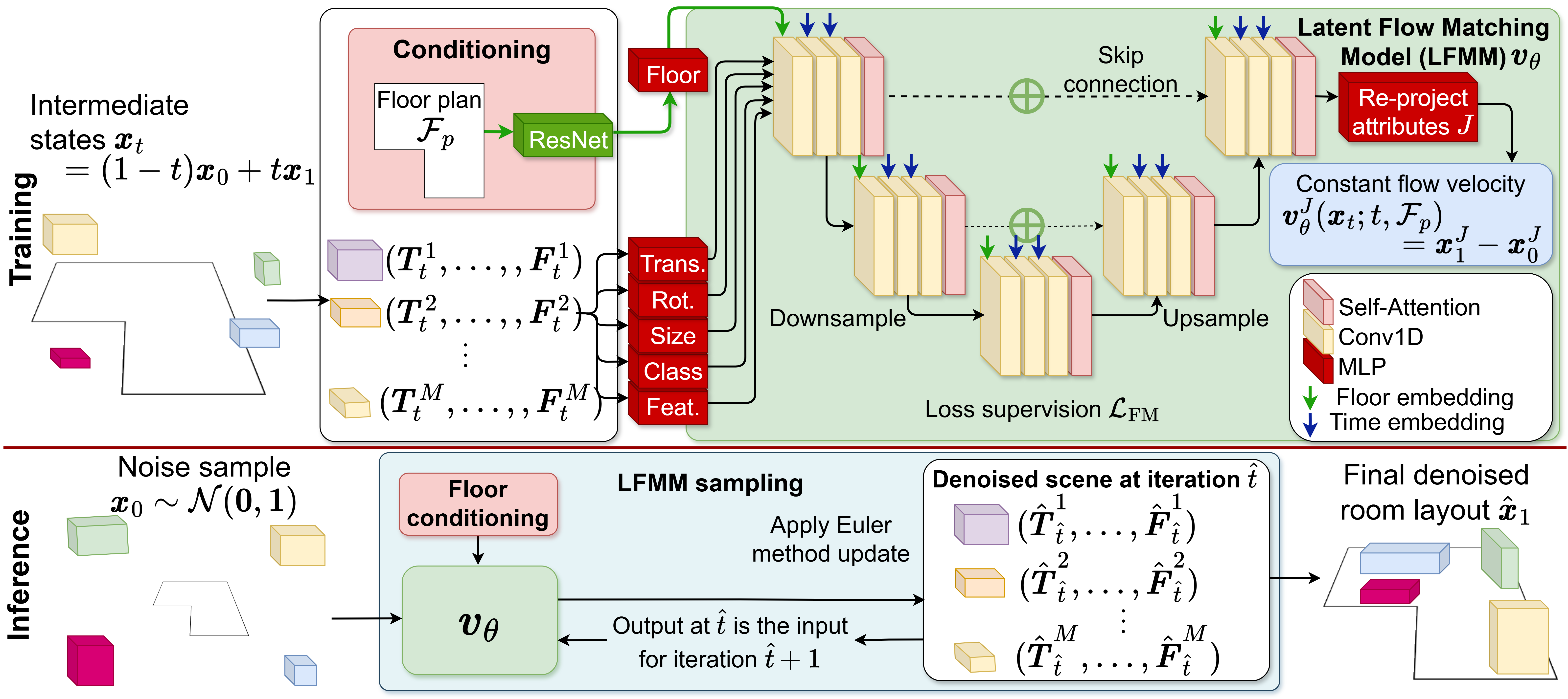}
    \caption{U-Net architecture of the Latent space Flow Matching Model.}
    \label{fig:lffm}
\end{figure*}

By contrast, the CPVQ-VAE partitions the codebook by class and the number of total codevectors becomes $N_K = N_c \times N_q$. Here, $N_c$ is the number of classes and $N_q$ is the number of codevectors assigned to each class. The quantization process is modified in the following way:
\begin{align}
    \mathcal{Q}(\mathcal{E}(\pmb{P}); \mathcal{C}, c)& = k^*_c =  \argmin_{k\in[N_K]}\norm{\mathcal{E}(\pmb{P}) - \mathbf{1}(c, k)\times\pmb{e}^k}^2_2,
    \label{eq:class-partitioned-quantization-process}
\end{align}
where the indicator function $\mathbf{1}(c, k)$ is used to partition the codebook $\mathcal{C}$ by object class $c\in[N_c]$. It is given by,
\begin{equation}
  \mathbf{1}(c, k):=
  \begin{cases}
      1 ~ &{\text{ if }} ~ (c-1)N_q \leq k < cN_q, \\
      0 ~ & \text{otherwise.}
  \end{cases}
\end{equation}
The \emph{class-aware} quantized feature is now $\pmb{z}^{q_c} = \pmb{e}^{k^*_c}$. Thereby, we ensure that codevectors $\pmb{z}^{q_c}$ belong to the appropriate class, $c$. The CPVQ-VAE is trained with the standard training objective:
\begin{align}
    \mathcal{L}_{\text{AE}} = \lambda_{\text{CD}}\mathcal{L}_{\text{CD}}(\pmb{P}, \tilde{\pmb{P}}) &+ \left\|\text{sg}(\mathcal{E}(\pmb{P}))-\pmb{z}^{q_c}\right\|^2_2 \nonumber \\ 
    &+ \left\|\mathcal{E}(\pmb{P})-\text{sg}(\pmb{z}^{q_c})\right\|^2_2,
    \label{eq:vqvae-loss}
\end{align}
where $\mathcal{L}_{\text{CD}}$ is the reconstruction loss given by the Chamfer distance between the input, $\pmb{P}$, and its reconstruction, $\tilde{\pmb{P}}$. The weight $\lambda_{\text{CD}}=10$ ensures all loss terms are of the same order and $\text{sg}(\cdot)$ represents the stop-gradient operator. 

\subsection{Class-Aware Running Average Update}
The training objective in Eq.~\eqref{eq:vqvae-loss} is susceptible to codebook collapse, a common problem in VQ-VAE architectures. At the start of training, the network optimizes a few codevectors within the codebook while a large number of codevectors are ignored. Codevectors which are favored for optimization at earlier times during training remain \emph{active} during training while those codevectors which do not receive substantial updates end up as \emph{dead} codevectors. This further leads to a degradation in the VQ-VAE's decoding ability as fewer viable codevectors result in fewer learned point cloud shapes.

To mitigate this issue, we reinitialize dead codevectors dynamically during training. Our approach is inspired by \citet{Zheng-Online-Clustered-Codebook}, however, we introduce a running average update strategy that reinitializes dead codevectors in a \emph{class-aware} manner whereas \citet{Zheng-Online-Clustered-Codebook} do not label codevectors by class and their proposed reinitialization is also class-agnostic. To achieve our dynamic, class-aware, reinitialization we first track average usage of codevectors, $U^k_s$, during each training step:
\begin{equation}
    U^k_s = \gamma U^k_{s-1} + \tfrac{1-\gamma}{B}u^k_s,
\end{equation}
where $s$ is the training step, $u^k_s$ is the number of encodings in the mini-batch which map to the codevector $\pmb{e}^k$, $U^k_0 = 0$, $B$ is the batch size, and $\gamma=0.99$ is a decay hyperparameter. Next, we select the closest features $\mathcal{E}^{i^*_c}(\pmb{P})$ in the mini-batch, to the codevectors $\pmb{e}^k$. These features become the anchors for the running average update, where we have,
\begin{align}
    i^*_c =  \argmin_{i\in[B]}\norm{\mathcal{E}^i(\pmb{P}) - \mathbf{1}(c, k)\times\pmb{e}^k}^2_2,
    \label{eq:class-aware-running-average-update}
\end{align}
with the index $i$ iterating over the entries of the mini-batch. Unlike previous works, we ensure that the 
method of 
anchor selection is \emph{class-aware} by using the indicator function to partition the codebook by class. Now, the anchor used to update the codevector $\pmb{e}^k$ is given by $\mathcal{E}^k(\pmb{P}) =\mathcal{E}^{i^*_c}(\pmb{P})$.

Finally, we ensure that dead codevectors are reinitialized while active ones remain unchanged. To do so, we reinitialize each codevector $\pmb{e}^k_s$ based on a decay value $\alpha^k_s$. We calculate $\alpha^k_s$ and update the codevector in the following manner:
\begin{align}
    \alpha^k_s &= \exp\left(-\tfrac{10U^k_s N_q}{1-\gamma} -\epsilon\right), \\
    \pmb{e}^k_s &= (1 - \alpha^k_s)\pmb{e}^k_{s-1} +\alpha^k_s \mathcal{E}^k(\pmb{P}).
\end{align}
This \emph{class-aware} running average update improves the CPVQ-VAE's learning ability as unused codevectors are more frequently reinitialized, which allows them to be better optimized during training. 

Given the trained CPVQ-VAE, we generate quantized latents $\pmb{z}^{q_c}$ of point cloud objects which are used for training the latent flow model described in Sec. \ref{sec:latent-flow}. As $\pmb{z}^{q_c}$ is a 128 dimensional vector, we truncate it to only the first 32 entries to form the object latent such that $\pmb{F}^m=\pmb{z}^{q_c}_{n<32}$. We found that using the 32 dimensional truncated latent was more efficient for training.

\subsection{Flow Matching on Optimal Transport Paths}
\label{sec:latent-flow}
Next, we introduce flow matching in latent space to evolve noisy inputs towards clean scene arrangements $\pmb{X}=\left\{\pmb{x}^1,\pmb{x}^2,\ldots,\pmb{x}^M, \mathcal{F}_p\right\}$ with $M$ objects and floor plan $\mathcal{F}_p$. The object attribute vector, with dimensionality $D_x$, is formed by concatenating all object parameters such that $\pmb{x}^m=(\pmb{T}^m; \pmb{R}^m; \pmb{S}^m; \pmb{C}^m; \pmb{F}^m)\in\mathbb{R}^{D_x}$. The sampling objective of generative methods can be expressed as a transport problem where samples from a noise distribution $p_0$, given by $\pmb{x}^m_0\sim p_0$, at time $t=0$, are transported towards clean samples in the data distribution $\pmb{x}^m_1\sim p_1(\pmb{x}|\mathcal{F}_p)$ at time $t=1$ where $p_1 = p_{\text{data}}$. For simplicity, we remove the superscript $m$ signifying the $m$-th object of the scene as the flow matching process is identical for all objects and only keep the subscript $t$ for time spent along the denoising trajectory.

In particular, we choose $p_0=\mathcal{N}(\pmb{0},\mathbf{1})$ as a simple starting distribution. For an optimal transport-based flow matching model, we aim to infer constant paths between samples, leading to smaller discretization errors and faster sampling. This can be accomplished if each intermediate state $\pmb{x}_t$ at time $t$ is a linear interpolation between $\pmb{x}_0$ and $\pmb{x}_1$,
\begin{equation}
\label{eq:linear-interpolation}
\pmb{x}_t = (1-t)\pmb{x}_0 + t\pmb{x}_1.
\end{equation}
Thereby, we deduce the velocity $\pmb{v}(\pmb{x}_t, t)$ to be constant for the above linearly interpolated states:
\begin{equation}
\label{eq:ode}
\frac{\text{d}\pmb{x}_t}{\text{d}t} = \pmb{v}(\pmb{x}_t; t, \mathcal{F}_p) = \pmb{x}_1-\pmb{x}_0,
\end{equation}
A constant flow velocity can be obtained by optimizing the following expected value:
\begin{equation}
\label{eq:ode-solution}
\min_{v}\int^1_0\mathbb{E}_{t\sim\mathcal{U}(0,1)}\left[\norm{\pmb{v}(\pmb{x}_t; t, \mathcal{F}_p) - (\pmb{x}_1 - \pmb{x}_0)}^2_2\right]\text{d}t,
\end{equation}
This flow velocity can be approximated by a deep neural network $\pmb{v}_\theta$ with network parameters $\theta$. Such a network aims to infer a constant flow $\pmb{v}_\theta(\pmb{x}_t; t, \mathcal{F}_p) = \pmb{x}_1 - \pmb{x}_0$ at each interpolated state $\pmb{x}_t$. We train this neural network using the following training objective:
\begin{equation}
\label{eq:velocity-flow-los}
\mathcal{L}_\text{FM} = \mathbb{E}_{t\sim\mathcal{U}(0,1)}\left[\lambda_J \sum_J \left( \norm{\pmb{v}^J_\theta(\pmb{x}_t; t, \mathcal{F}_p) - (\pmb{x}^J_1 - \pmb{x}^J_0)}^2_2\right)\right],
\end{equation}
where the $\lambda_J$ is a weight that accounts for the loss contribution of each object parameter type, $J=T, R, S, C, F$, and $\pmb{v}^J_\theta(\pmb{x}_t; t, \mathcal{F}_p)$ are the entries of  $\pmb{v}_\theta(\pmb{x}_t; t, \mathcal{F}_p)$ that correspond to the $J$-th parameter. Fig. \ref{fig:lffm} provides an overview of the our latent flow matching model.

\subsection{Velocity Flow Sampling}
A trained velocity flow model can be used to generate plausible scene layouts using the Euler method for sampling, as it provides a numerical solution to Eq.~\eqref{eq:ode}. Consequently, the denoised object parameters, at each iteration, are given by $\hat{\pmb{x}}_{(\hat{t}+1)/N_{\hat{t}}} = \hat{\pmb{x}}_{\hat{t}/N_{\hat{t}}} + \frac{1}{N_{\hat{t}}}\pmb{v}_\theta(\hat{\pmb{x}}_{\hat{t}/N_{\hat{t}}}; \hat{t}/N_{\hat{t}}, \mathcal{F}_p)$ where $\hat{t} = \left\{0, 1, 2, \ldots, N_{\hat{t}}\right\}$ are integer time steps and $N_{\hat{t}}=100$ is the total number of steps.

\midsepremove
\setlength{\tabcolsep}{1mm}
\begin{table*}[!t]
\small
\renewcommand\arraystretch{1.2}
\begin{center}
    \begin{tabular}{*{13}{l|cccc|cccc|cccc}}
        \toprule
        \multirow{2}*{Generation Method} & \multicolumn{4}{c|}{Living room}   & \multicolumn{4}{c|}{Dining room}  & \multicolumn{4}{c}{Bedroom} \\
            \cmidrule(lr){2-13}
             & \multicolumn{2}{c}{CD $\downarrow$} & \multicolumn{2}{c|}{P2M $\downarrow$} &
             \multicolumn{2}{c}{CD $\downarrow$} & \multicolumn{2}{c|}{P2M $\downarrow$} & 
             \multicolumn{2}{c}{CD $\downarrow$} & \multicolumn{2}{c}{P2M $\downarrow$} \\ %
            \midrule
            Diffuscene & \multicolumn{2}{c}{30.63} & \multicolumn{2}{c|}{29.87} 
            & \multicolumn{2}{c}{30.60} & \multicolumn{2}{c|}{29.49} 
            & \multicolumn{2}{c}{45.01} & \multicolumn{2}{c}{44.88} \\
            Ours (LFMM + VAE) &  \multicolumn{2}{c}{\underline{24.65}} & \multicolumn{2}{c|}{\underline{23.41}} 
            &  \multicolumn{2}{c}{\underline{2.66}} & \multicolumn{2}{c|}{\underline{2.62}}
            &  \multicolumn{2}{c}{\underline{4.24}} & \multicolumn{2}{c}{\underline{3.63}}  \\
            Ours (LFMM + CPVQ-VAE) & \multicolumn{2}{c}{\textbf{9.06}} & \multicolumn{2}{c|}{\textbf{8.27}} 
            & \multicolumn{2}{c}{\textbf{2.38}} & \multicolumn{2}{c|}{\textbf{2.17}} 
            & \multicolumn{2}{c}{\textbf{2.46}} & \multicolumn{2}{c}{\textbf{2.06}}\\
        \bottomrule
        Retrieval Method & FID $\downarrow$ & KID $\downarrow$ & SCA $\%$ & CKL $\downarrow$   & FID $\downarrow$ & KID $\downarrow$ & SCA $\%$ & CKL $\downarrow$ & FID $\downarrow$ & KID $\downarrow$ & SCA  $\%$ & CKL $\downarrow$ \\ 
        
        \midrule

        ATISS
              & 27.78 & 5.869 & 0.617 & \underline{0.560} 
              & 27.82 & 3.470 & 0.557 & \underline{0.633} 
              & \underline{23.02} & 1.720 & \textbf{0.498} & \textbf{2.568} \\

        Diffuscene
              & 27.39 & 6.169 & 0.615 & \textbf{0.455}
              & 30.23 & 5.697 & 0.562 & \textbf{0.546}
              & 25.13 & 2.459 & 0.606 & 6.184 \\

        Ours (LFMM + VAE)
              & \textbf{24.46} & \textbf{4.020} & \textbf{0.608} & 1.063 
              & \underline{25.89} & \underline{2.945} & \underline{0.552} & 1.309
              & 21.86 & \textbf{0.723} & 0.545 & 3.095 \\
              
        Ours (LFMM + CPVQ-VAE)
              & \underline{26.10} & \underline{5.818} & \underline{0.614} & 1.129
              & \textbf{25.58} & \textbf{2.708} & \textbf{0.484} & 1.176
              & \textbf{21.34} & \underline{}{0.794} & \underline{0.534} & \underline{3.837} \\

    \bottomrule
    \end{tabular}
    \caption{Comparison on point cloud generation and scene retrieval. Top results are in bold while second best are underlined. CD, P2M and KID results are multiplied by $10^3$ and CKL by $10^2$. SCA scores closer to $0.5$ are better}
    \label{tab:room-cond}
    \end{center}
\end{table*}
\midsepdefault

\begin{figure*}[!t]
    \centering
    \includegraphics[width=\linewidth]{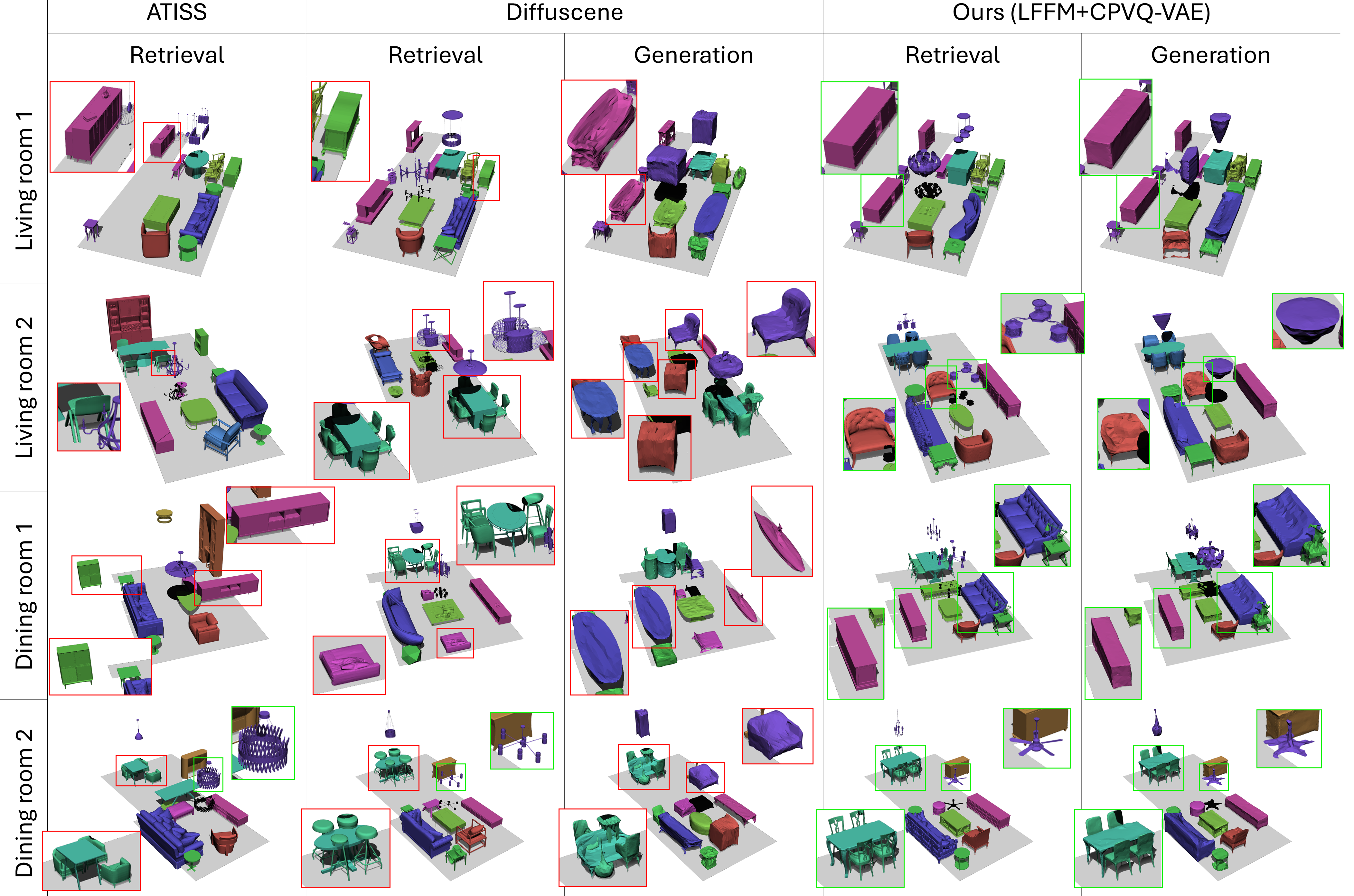}
    \caption{Visualization of retrieved and generated scenes. Green close-ups indicate correctly retrieved/decoded objects while incorrect results are in red. Notably, unlike the VAE decoder used by Diffuscene, our CPVQ-VAE decoder accurately decodes object features into class-consistent point clouds.
    }
    \label{fig:results}
\end{figure*}

\subsection{Class-Aware Inverse Look-up}
The denoised feature $\hat{\pmb{F}}$ is a $32$ dimensional vector while codebook entries $\pmb{e}^k$ are $128$ dimensional. We must look-up the correct codebook entry related to $\hat{\pmb{F}}$ which also agrees with the generated class $\hat{c}=\argmax_{n\in [N_c]}\hat{\pmb{C}}_n$ (where $\hat{\pmb{C}}_n$ is the $n$-th element of $\hat{\pmb{C}}\in\mathbb{R}^{N_c}$). Therefore, we propose the look-up function $\mathbf{L}$ w.r.t truncated codevectors $\pmb{e}^k_{n<32}$ (consisting of the first 32 elements of $\pmb{e}^k$),
\begin{align}
    \mathbf{L}(\hat{\pmb{F}}, \mathcal{C}, \hat{c}) = k^*_{\hat{c}} =  \argmax_{k\in[N_K]}\left(\hat{\pmb{F}} \cdot \left[\mathbf{1}(\hat{c}, k)\times\pmb{e}^k_{n<32}\right]\right),
    \label{eq:flow-feature-to-quantized-feature-inverse-look-up}
\end{align}
that maximizes the cosine similarity between $\pmb{e}^k_{n<32}$ and denoised object features, in a class-aware manner. 

Finally we recover the quantized feature $\pmb{z}^{q_{\hat{c}}} = \pmb{e}^{k^*_{\hat{c}}}$, which is subsequently decoded by the CPVQ-VAE to give the generated point cloud $\tilde{\pmb{P}} = \mathcal{D}(\pmb{z}^{q_{\hat{c}}})$.

\section{Experiments}
We follow the work of \citet{Paschalidou-ATISS} and train and test our models on the 3D-FRONT dataset \cite{Fu-3D-FRONT, Fu-3D-FUTURE} containing living, dining and bed rooms. Specifically, we consider the version of the 3D-FRONT used by \citet{Wei-LEGO-Net} and \citet{Maillard-Debara}. The pre-processing of ATISS \cite{Paschalidou-ATISS} gives training and testing splits of 2338/587, 2071/516 and 5668/224 for living room, dining room and bedroom scenes, respectively. Additional details are provided in the \emph{supplementary}. 

\subsubsection{Implementation}
Separate networks are trained on the living and dining room training sets for 100K epochs and on the bedroom training set for 40K epochs, respectively. Training and inference is performed on an NVIDIA RTX 4090 graphics card, using PyTorch for the implementation code. 

\subsubsection{Benchmark Comparisons}
We compare generation results with Diffuscene \cite{Tang-Diffuscene}, and a variant of our LFMM which uses the same pre-trained VAE as Diffuscene. For retrieval, we additionally compare with ATISS \cite{Paschalidou-ATISS}. We do not compare with DeBaRa \cite{Maillard-Debara} or CASAGPT \cite{Feng-CASAGPT} as no code was available, and all generation methods are retrained to ensure a fair comparison across all settings. For visualizing denoised point clouds, we employ alpha wrapping to reconstruct meshes \cite{Edelsbrunner-Alpha-Shapes}. For point cloud generation, decoded point clouds are evaluated w.r.t their counterparts retrieved from the predefined objects database. We treat these retrieved objects as the ground truth and compute Chamfer Distance (CD $\times10^3$) and Point2Mesh Distance (P2M $\times10^3$) metrics using PyTorch3D \cite{Ravi-PyTorch3D}. To evaluate retrieval results, Fr\'{e}chet Inception Distance (FID), Kernel Inception Distance (KID $\times10^3$), Scene Classification Accuracy and Categorical KL-Divergence (CKL $\times10^2$) are used. The \emph{supplementary document} contains more details regarding comparisons.

\subsection{Point Cloud Generation}
The top half of Table \ref{tab:room-cond} and the \emph{generation} columns of Fig. \ref{fig:results} indicate the performance of different methods on the point cloud generation task. We observe that Diffuscene regularly decodes incorrect point cloud objects which are not coherent in class labels and bounding box parameters on both living and dining room sets (see red close-ups in Fig. \ref{fig:results}). Table \ref{tab:room-cond} shows that our LFMM with a standard VAE decoder suffers from the same drawback on more complex living room scenes while producing much better object features (w.r.t Diffuscene) for dining room scenes. By contrast, our LFMM together with the CPVQ-VAE, successfully decodes point cloud objects that agree with their target class labels across all scene types. Notably, on complex living room data, we achieve up to 70.4\% and 72.3\% reduction in CD and P2M errors, compared to Diffuscene. With respect to our LFMM+VAE variant, our full model showcases a 63.2\% and 64.7\% reduction in CD and P2M errors, respectively. 

\subsection{Point Cloud Retrieval}
The bottom half of Table \ref{tab:room-cond} and the \emph{retrieval} columns of Fig. \ref{fig:results} demonstrate each method's retrieval performance. In general, we see that retrieval using flow matching is superior to scenes retrieved via diffusion (Diffuscene) or autoregression (ATISS). Although ATISS has a simpler training objective, which does not require the network to regress latent features, its performance on retrieval metrics is sub-optimal compared to our LFMM variants. Moreover, both ATISS, and Diffuscene, frequently generate incorrect bounding box parameters which lead to object collisions or object placements beyond the floor plan (see Fig. \ref{fig:results} for visualizations).

\midsepremove
\begin{table}[!t]
\small
\renewcommand\arraystretch{1.2}
\begin{center}
    \begin{tabular}{l|ccc}
        \toprule
            Method & Runtime (s)  $\downarrow$ & Avg. FID  $\downarrow$ & Avg. KID  $\downarrow$ \\
            \midrule              
            ATISS & \textbf{0.024} & \underline{26.21} & \underline{3.686} \\
            Diffuscene & 9.153 & 27.58 & 4.775 \\
            Ours & \underline{0.892} & \textbf{24.34} & \textbf{3.107}\\
        \bottomrule
    \end{tabular}
    \caption{Runtimes of different methods along with average FID and KID performance across all scene types.}
    \label{tab:runtime}
    \end{center}
\end{table}
\midsepdefault

\paragraph{Runtime efficiency}
Table \ref{tab:runtime} provides runtime and average performance results. ATISS is the most efficient among all methods but remains sub-optimal in retrieval. Furthermore, among methods which generate the complete set of object attributes, our method has a clear advantage in efficiency. Concretely, our method is $90.3\%$ faster than Diffuscene.

\section{Ablations}
To showcase the effectiveness of our proposed contributions, we investigate the following: (1) the role of autoencoder design in point cloud generation and (2) the impact of sampling iteration number on scene retrieval. For (1), we propose four autoencoder variants (V1 to V4) and (2), we consider our preferred CPVQ-VAE+LFMM model with different sampling iteration numbers $N_{\hat{t}}$.

\paragraph{Autoencoder design for point cloud generation.} Table \ref{tab:ablation-vqvae} shows our full CPVQ-VAE (V4) significantly improves upon the traditional VAE (V1), with a $41.8\%$ and $43.1\%$ reduction in CD and P2M error. On more complex scenes such as living rooms, the gap in CD and P2M results are wider as evidenced by table \ref{tab:room-cond}. We also see significant differences in CD and P2M results between a simple VQ-VAE (V2) and a class-partitioned VQ-VAE (V3), as class partitioning enforces class-consistent decoding of latents. However, both (V2) and (V3) suffer from the issue of codebook collapse and, in turn, perform worse than a standard VAE (V1). The full CPVQ-VAE (V4) reinforces the importance of partitioning the codebook by class \emph{and} applying the class-aware running average update to reinitialize dead codevectors during training, which leads to impressive gains over (V1). 

\paragraph{Sampling iteration number and scene retrieval.} While the CPVQ-VAE+LFMM model infers plausible scene arrangements at low $N_{\hat{t}}$ (table \ref{tab:ablation-lffm}), it produces sub-optimal performance. In fact, a higher number of sampling iterations is necessary to obtain consistent bounding box parameters to correctly position scene objects. However, this comes at the price of higher latency. We find $N_{\hat{t}}=100$ is the optimal number of sampling iterations that balances performance and speed when generating object bounding box parameters.

\midsepremove
\begin{table}[!t]
    \small
    \renewcommand\arraystretch{1.2}
    \begin{center}
        \begin{tabular}{l|cccc|cc}
            \toprule
            \multirow{2}*{Variant} & \multicolumn{4}{c|}{Module} & \multicolumn{2}{c}{Bedroom} \\
                \cmidrule(lr){2-7}
                & VAE & VQ-VAE & CP & RAU & CD $\downarrow$ & P2M $\downarrow$ \\  
                \midrule
                V1 & \checkmark & - & - & - & \underline{4.24} & \underline{3.63} \\
                V2 & - & \checkmark & - & - & 36.27 & 33.93 \\
                V3 & - & \checkmark & \checkmark & - & 5.00 & 4.17 \\
                V4 & - & \checkmark & \checkmark & \checkmark & \textbf{2.46} & \textbf{2.06}  \\
            \bottomrule
        \end{tabular}
        \caption{Ablation on autoencoder variants. CP denotes class partitioning and RAU is the running average update.}
    \label{tab:ablation-vqvae}
    \end{center}
\end{table}
\midsepdefault

\midsepremove
\begin{table}[!t]
\small
\renewcommand\arraystretch{1.2}
\begin{center}
    \begin{tabular}{*{6}{l|ccccc}}
        \toprule
            \multirow{2}*{$N_{\hat{t}}$} & \multicolumn{5}{c}{Bedroom} \\
        \cmidrule(lr){2-6}
            & FID $\downarrow$ & KID $\downarrow$ & SCA  $\%$ & CKL $\downarrow$ & Runtime (s) $\downarrow$ \\
        \midrule              
            10 & 21.79 & 1.324 & 0.537 & 3.954 & 0.094 \\
            100 & \textbf{21.34} & \textbf{0.794} & \underline{0.534} & \underline{3.837} & 0.892 \\
            1000 & \underline{21.75} & \underline{0.967} & \textbf{0.517} & \textbf{2.411} & 8.963 \\
        \bottomrule
    \end{tabular}
    \caption{Ablation results on flow matching iterations $N_{\hat{t}}$ along with respective runtimes in seconds.}
    \label{tab:ablation-lffm}
    \end{center}
\end{table}
\midsepdefault

\section{Limitations and Future Work}
Our method makes significant contributions to point cloud scene generation. However, current autoencoders cannot process dense point clouds and we limit ourselves to objects of $N_P = 2025$ points. Generating denser point clouds would yield better mesh reconstructions but requires improving the learning capacity of autoencoders. Furthermore, vector quantization introduces quantization error which marginally deteriorates the variety of \emph{retrieved} objects. Therefore, scaling the CPVQ-VAE learning capacity and reducing quantization error are directions for future research.

\section{Conclusion}
In this paper, we introduced the first scene-level point cloud generation mechanism, capable of handling complex scenes of varied, multi-categorical objects. We introduced a class-partitioned vector quantized variational autoencoder (CPVQ-VAE) with labeled codebook entries that, given class labels and using a class-aware inverse look-up, successfully decodes object latents into the correct point cloud shapes which agree with target classes. We alleviate codebook collapse within the CPVQ-VAE via a class-aware running average update mechanism. Both object latents and class labels, along with object bounding box parameters, are generated by a latent space flow matching model designed specifically to handle scene-level generation. Extensive experiments across multiple room types reveal that our method reliably recovers point cloud scenes with consistent object properties, unlike previous methods, and promotes a new direction for future 3D scene generation.

\section*{Acknowledgments}
This research was supported by the Australian Government through the Australian Research Council's Discovery Projects funding scheme (project DP240101926).

\bibliography{aaai2026}

\begin{thebibliography}{54}
\providecommand{\natexlab}[1]{#1}

\bibitem[{Albergo and Vanden{-}Eijnden(2023)}]{Albergo-Stochastic-Interpolants}
Albergo, M.~S.; and Vanden{-}Eijnden, E. 2023.
\newblock Building Normalizing Flows with Stochastic Interpolants.
\newblock In \emph{the 11th International Conference on Learning Representations (ICLR)}.

\bibitem[{Cai et~al.(2020)Cai, Yang, Averbuch-Elor, Hao, Belongie, Snavely, and Hariharan}]{Cai-ShapeGF}
Cai, R.; Yang, G.; Averbuch-Elor, H.; Hao, Z.; Belongie, S.; Snavely, N.; and Hariharan, B. 2020.
\newblock Learning Gradient Fields for Shape Generation.
\newblock In \emph{Computer Vision – ECCV 2020}, 364--381.

\bibitem[{Chen et~al.(2018)Chen, Rubanova, Bettencourt, and Duvenaud}]{Chen-Neural-ODE}
Chen, T.~Q.; Rubanova, Y.; Bettencourt, J.; and Duvenaud, D. 2018.
\newblock Neural Ordinary Differential Equations.
\newblock In \emph{Advances in Neural Information Processing Systems (NeurIPS)}, 6572--6583.

\bibitem[{Chorowski et~al.(2019)Chorowski, Weiss, Bengio, and van~den Oord}]{Chorowski-Wavenet}
Chorowski, J.; Weiss, R.~J.; Bengio, S.; and van~den Oord, A. 2019.
\newblock Unsupervised Speech Representation Learning Using WaveNet Autoencoders.
\newblock \emph{{IEEE} {ACM} Trans. Audio Speech Lang. Process.}, 27(12): 2041--2053.

\bibitem[{de~Silva~Edirimuni et~al.(2024)de~Silva~Edirimuni, Lu, Li, Wei, Robles{-}Kelly, and Li}]{Edirimuni-StraightPCF}
de~Silva~Edirimuni, D.; Lu, X.; Li, G.; Wei, L.; Robles{-}Kelly, A.; and Li, H. 2024.
\newblock StraightPCF: Straight Point Cloud Filtering.
\newblock In \emph{Proceedings of the Computer Vision and Pattern Recognition Conference (CVPR)}, 20721--20730.

\bibitem[{de~Silva~Edirimuni et~al.(2023)de~Silva~Edirimuni, Lu, Shao, Li, Robles{-}Kelly, and He}]{Edirimuni-IterativePFN}
de~Silva~Edirimuni, D.; Lu, X.; Shao, Z.; Li, G.; Robles{-}Kelly, A.; and He, Y. 2023.
\newblock IterativePFN: True Iterative Point Cloud Filtering.
\newblock In \emph{Proceedings of the Computer Vision and Pattern Recognition Conference (CVPR)}, 13530--13539.

\bibitem[{Edelsbrunner and M{\"{u}}cke(1994)}]{Edelsbrunner-Alpha-Shapes}
Edelsbrunner, H.; and M{\"{u}}cke, E.~P. 1994.
\newblock Three-dimensional alpha shapes.
\newblock \emph{{ACM} Trans. Graph.}, 13(1): 43--72.

\bibitem[{Feng et~al.(2025)Feng, Zhou, Liao, Cheng, and Zhou}]{Feng-CASAGPT}
Feng, W.; Zhou, H.; Liao, J.; Cheng, L.; and Zhou, W. 2025.
\newblock {CASAGPT:} Cuboid Arrangement and Scene Assembly for Interior Design.
\newblock In \emph{Proceedings of the Computer Vision and Pattern Recognition Conference (CVPR)}, 29173--29182.

\bibitem[{Fu et~al.(2021{\natexlab{a}})Fu, Cai, Gao, Zhang, Wang, Li, Zeng, Sun, Jia, Zhao et~al.}]{Fu-3D-FRONT}
Fu, H.; Cai, B.; Gao, L.; Zhang, L.-X.; Wang, J.; Li, C.; Zeng, Q.; Sun, C.; Jia, R.; Zhao, B.; et~al. 2021{\natexlab{a}}.
\newblock 3d-front: 3d furnished rooms with layouts and semantics.
\newblock In \emph{Proceedings of the IEEE/CVF International Conference on Computer Vision}, 10933--10942.

\bibitem[{Fu et~al.(2021{\natexlab{b}})Fu, Jia, Gao, Gong, Zhao, Maybank, and Tao}]{Fu-3D-FUTURE}
Fu, H.; Jia, R.; Gao, L.; Gong, M.; Zhao, B.; Maybank, S.; and Tao, D. 2021{\natexlab{b}}.
\newblock 3d-future: 3d furniture shape with texture.
\newblock \emph{International Journal of Computer Vision}, 1--25.

\bibitem[{Hinton and Salakhutdinov(2006)}]{Hinton-Dimensionality-Reduction}
Hinton, G.~E.; and Salakhutdinov, R.~R. 2006.
\newblock Reducing the Dimensionality of Data with Neural Networks.
\newblock \emph{Science}, 313(5786): 504--507.

\bibitem[{Hinton and Zemel(1993)}]{Hinton-Autoencoder}
Hinton, G.~E.; and Zemel, R. 1993.
\newblock Autoencoders, Minimum Description Length and Helmholtz Free Energy.
\newblock In \emph{Advances in Neural Information Processing Systems (NIPS)}, volume~6.

\bibitem[{Ho, Jain, and Abbeel(2020)}]{Ho-DDPM}
Ho, J.; Jain, A.; and Abbeel, P. 2020.
\newblock Denoising Diffusion Probabilistic Models.
\newblock In \emph{Advances in Neural Information Processing Systems (NeurIPS)}.

\bibitem[{Hyvärinen(2005)}]{Hyvarinen-Score}
Hyvärinen, A. 2005.
\newblock Estimation of Non-Normalized Statistical Models by Score Matching.
\newblock \emph{Journal of Machine Learning Research}, 6: 695--709.

\bibitem[{Ju et~al.(2024)Ju, Huang, Li, Zhang, Qiao, and Li}]{Ju-DiffInDScene}
Ju, X.; Huang, Z.; Li, Y.; Zhang, G.; Qiao, Y.; and Li, H. 2024.
\newblock DiffInDScene: Diffusion-Based High-Quality 3D Indoor Scene Generation.
\newblock In \emph{Proceedings of the Computer Vision and Pattern Recognition Conference (CVPR)}, 4526--4535.

\bibitem[{Kingma et~al.(2014)Kingma, Mohamed, Rezende, and Welling}]{Kingma-Generalized-Scalable-VAE}
Kingma, D.~P.; Mohamed, S.; Rezende, D.~J.; and Welling, M. 2014.
\newblock Semi-supervised Learning with Deep Generative Models.
\newblock In \emph{Advances in Neural Information Processing Systems (NIPS)}, 3581--3589.

\bibitem[{Kingma and Welling(2014)}]{Kingma-VAE}
Kingma, D.~P.; and Welling, M. 2014.
\newblock Auto-Encoding Variational Bayes.
\newblock In \emph{the 2nd International Conference on Learning Representations (ICLR)}.

\bibitem[{Lee et~al.(2022)Lee, Kim, Kim, Cho, and Han}]{Lee-Residual-Quantization}
Lee, D.; Kim, C.; Kim, S.; Cho, M.; and Han, W. 2022.
\newblock Autoregressive Image Generation using Residual Quantization.
\newblock In \emph{Proceedings of the Computer Vision and Pattern Recognition Conference (CVPR)}, 11513--11522.

\bibitem[{Li et~al.(2025)Li, Zou, Liu, Wang, Liang, Yu, Liu, Guo, Liang, Ouyang, and Cao}]{Li-TripoSG}
Li, Y.; Zou, Z.; Liu, Z.; Wang, D.; Liang, Y.; Yu, Z.; Liu, X.; Guo, Y.; Liang, D.; Ouyang, W.; and Cao, Y. 2025.
\newblock TripoSG: High-Fidelity 3D Shape Synthesis using Large-Scale Rectified Flow Models.
\newblock \emph{CoRR}, abs/2502.06608.

\bibitem[{Lipman et~al.(2023)Lipman, Chen, Ben{-}Hamu, Nickel, and Le}]{Lipman-Flow-Matching}
Lipman, Y.; Chen, R. T.~Q.; Ben{-}Hamu, H.; Nickel, M.; and Le, M. 2023.
\newblock Flow Matching for Generative Modeling.
\newblock In \emph{the 11th International Conference on Learning Representations (ICLR)}.

\bibitem[{Liu, Gong, and Liu(2023)}]{Liu-RectifiedFlow}
Liu, X.; Gong, C.; and Liu, Q. 2023.
\newblock Flow Straight and Fast: Learning to Generate and Transfer Data with Rectified Flow.
\newblock In \emph{the 11th International Conference on Learning Representations (ICLR)}.

\bibitem[{Luo and Hu(2021{\natexlab{a}})}]{Luo-Diffusive}
Luo, S.; and Hu, W. 2021{\natexlab{a}}.
\newblock Diffusion Probabilistic Models for 3D Point Cloud Generation.
\newblock In \emph{Proceedings of the IEEE/CVF Conference on Computer Vision and Pattern Recognition (CVPR)}, 2837--2845.

\bibitem[{Luo and Hu(2021{\natexlab{b}})}]{Luo-Score-Based-Denoising}
Luo, S.; and Hu, W. 2021{\natexlab{b}}.
\newblock Score-Based Point Cloud Denoising.
\newblock In \emph{Proceedings of the IEEE/CVF International Conference on Computer Vision (ICCV)}, 4583--4592.

\bibitem[{Maillard et~al.(2024)Maillard, Sereyjol-Garros, Durand, and Ovsjanikov}]{Maillard-Debara}
Maillard, L.; Sereyjol-Garros, N.; Durand, T.; and Ovsjanikov, M. 2024.
\newblock DeBaRA: Denoising-Based 3D Room Arrangement Generation.
\newblock In \emph{Advances in Neural Information Processing Systems (NeurIPS)}, volume~37, 109202--109232.

\bibitem[{Maoutsa, Reich, and Opper(2020)}]{Maoutsa-Deterministic-LE}
Maoutsa, D.; Reich, S.; and Opper, M. 2020.
\newblock Interacting Particle Solutions of Fokker–Planck Equations Through Gradient–Log–Density Estimation.
\newblock \emph{Entropy}, 22.

\bibitem[{Paschalidou et~al.(2021)Paschalidou, Kar, Shugrina, Kreis, Geiger, and Fidler}]{Paschalidou-ATISS}
Paschalidou, D.; Kar, A.; Shugrina, M.; Kreis, K.; Geiger, A.; and Fidler, S. 2021.
\newblock ATISS: Autoregressive Transformers for Indoor Scene Synthesis.
\newblock In \emph{Advances in Neural Information Processing Systems (NeurIPS)}.

\bibitem[{Qi et~al.(2017{\natexlab{a}})Qi, Su, Mo, and Guibas}]{Qi-PointNet}
Qi, C.; Su, H.; Mo, K.; and Guibas, L. 2017{\natexlab{a}}.
\newblock PointNet: Deep Learning on Point Sets for 3D Classification and Segmentation.
\newblock In \emph{Proceedings of the Computer Vision and Pattern Recognition Conference (CVPR)}, 77--85.

\bibitem[{Qi et~al.(2017{\natexlab{b}})Qi, Yi, Su, and Guibas}]{Qi-PointNet++}
Qi, C.~R.; Yi, L.; Su, H.; and Guibas, L.~J. 2017{\natexlab{b}}.
\newblock PointNet++: Deep Hierarchical Feature Learning on Point Sets in a Metric Space.
\newblock In \emph{Advances in Neural Information Processing Systems (NeurIPS)}, volume~30.

\bibitem[{Radford et~al.(2018)Radford, Narasimhan, Salimans, Sutskever et~al.}]{Radford-ChatGPT}
Radford, A.; Narasimhan, K.; Salimans, T.; Sutskever, I.; et~al. 2018.
\newblock Improving language understanding by generative pre-training.
\newblock \emph{OpenAI Blog}.

\bibitem[{Ravi et~al.(2020)Ravi, Reizenstein, Novotny, Gordon, Lo, Johnson, and Gkioxari}]{Ravi-PyTorch3D}
Ravi, N.; Reizenstein, J.; Novotny, D.; Gordon, T.; Lo, W.-Y.; Johnson, J.; and Gkioxari, G. 2020.
\newblock Accelerating 3D Deep Learning with PyTorch3D.
\newblock \emph{arXiv preprint arXiv:2007.08501}.

\bibitem[{Razavi, van~den Oord, and Vinyals(2019)}]{Razavi-VQVAE2}
Razavi, A.; van~den Oord, A.; and Vinyals, O. 2019.
\newblock Generating Diverse High-Fidelity Images with {VQ-VAE-2}.
\newblock In \emph{Advances in Neural Information Processing Systems (NeurIPS)}, 14837--14847.

\bibitem[{Ritchie, Wang, and Lin(2019)}]{Ritchie-ISS-DCGM}
Ritchie, D.; Wang, K.; and Lin, Y. 2019.
\newblock Fast and Flexible Indoor Scene Synthesis via Deep Convolutional Generative Models.
\newblock In \emph{Proceedings of the Computer Vision and Pattern Recognition Conference (CVPR)}, 6182--6190.

\bibitem[{Rombach et~al.(2022)Rombach, Blattmann, Lorenz, Esser, and Ommer}]{Rombach-Stable-Diffusion}
Rombach, R.; Blattmann, A.; Lorenz, D.; Esser, P.; and Ommer, B. 2022.
\newblock High-Resolution Image Synthesis with Latent Diffusion Models.
\newblock In \emph{Proceedings of the Computer Vision and Pattern Recognition Conference (CVPR)}, 10674--10685.

\bibitem[{Ronneberger, Fischer, and Brox(2015)}]{Ronnenberger-U-Net}
Ronneberger, O.; Fischer, P.; and Brox, T. 2015.
\newblock U-Net: Convolutional Networks for Biomedical Image Segmentation.
\newblock In \emph{Medical Image Computing and Computer-Assisted Intervention -- MICCAI 2015}, volume 9351, 234--241.

\bibitem[{Salimans et~al.(2017)Salimans, Karpathy, Chen, and Kingma}]{Salimans-PixelCNN++}
Salimans, T.; Karpathy, A.; Chen, X.; and Kingma, D.~P. 2017.
\newblock PixelCNN++: Improving the PixelCNN with Discretized Logistic Mixture Likelihood and Other Modifications.
\newblock In \emph{the 5th International Conference on Learning Representations (ICLR)}.

\bibitem[{Song, Meng, and Ermon(2021)}]{Song-DDIM}
Song, J.; Meng, C.; and Ermon, S. 2021.
\newblock Denoising Diffusion Implicit Models.
\newblock In \emph{the 9th International Conference on Learning Representations (ICLR)}.

\bibitem[{Song et~al.(2023)Song, Dhariwal, Chen, and Sutskever}]{Song-Consistency}
Song, Y.; Dhariwal, P.; Chen, M.; and Sutskever, I. 2023.
\newblock Consistency models.
\newblock In \emph{Proceedings of the 40th International Conference on Machine Learning (ICML)}, 32211–32252.

\bibitem[{Song and Ermon(2019)}]{Song-Score-Matching}
Song, Y.; and Ermon, S. 2019.
\newblock Generative modeling by estimating gradients of the data distribution.
\newblock In \emph{Advances in Neural Information Processing Systems (NeurIPS)}.

\bibitem[{Song et~al.(2021)Song, Sohl-Dickstein, Kingma, Kumar, Ermon, and Poole}]{Song-SDE-PF-ODE}
Song, Y.; Sohl-Dickstein, J.; Kingma, D.~P.; Kumar, A.; Ermon, S.; and Poole, B. 2021.
\newblock Score-Based Generative Modeling through Stochastic Differential Equations.
\newblock In \emph{International Conference on Learning Representations}.

\bibitem[{Takida et~al.(2024)Takida, Ikemiya, Shibuya, Shimada, Choi, Lai, Murata, Uesaka, Uchida, Liao, and Mitsufuji}]{Takida-HQ-VAE}
Takida, Y.; Ikemiya, Y.; Shibuya, T.; Shimada, K.; Choi, W.; Lai, C.; Murata, N.; Uesaka, T.; Uchida, K.; Liao, W.; and Mitsufuji, Y. 2024.
\newblock {HQ-VAE:} Hierarchical Discrete Representation Learning with Variational Bayes.
\newblock \emph{Trans. Mach. Learn. Res.}, 2024.

\bibitem[{Takida et~al.(2022)Takida, Shibuya, Liao, Lai, Ohmura, Uesaka, Murata, Takahashi, Kumakura, and Mitsufuji}]{Takida-SQ-VAE}
Takida, Y.; Shibuya, T.; Liao, W.; Lai, C.; Ohmura, J.; Uesaka, T.; Murata, N.; Takahashi, S.; Kumakura, T.; and Mitsufuji, Y. 2022.
\newblock {SQ-VAE:} Variational Bayes on Discrete Representation with Self-annealed Stochastic Quantization.
\newblock In \emph{Proceedings of the 39th International Conference on Machine Learning (ICML)}, volume 162 of \emph{Proceedings of Machine Learning Research}, 20987--21012.

\bibitem[{Tang et~al.(2024)Tang, Nie, Markhasin, Dai, Thies, and Nießner}]{Tang-Diffuscene}
Tang, J.; Nie, Y.; Markhasin, L.; Dai, A.; Thies, J.; and Nießner, M. 2024.
\newblock DiffuScene: Denoising Diffusion Models for Generative Indoor Scene Synthesis.
\newblock In \emph{Proceedings of the Computer Vision and Pattern Recognition Conference (CVPR)}, 20507--20518.

\bibitem[{van~den Oord et~al.(2016{\natexlab{a}})van~den Oord, Dieleman, Zen, Simonyan, Vinyals, Graves, Kalchbrenner, Senior, and Kavukcuoglu}]{van-den-Oord-Wavenet}
van~den Oord, A.; Dieleman, S.; Zen, H.; Simonyan, K.; Vinyals, O.; Graves, A.; Kalchbrenner, N.; Senior, A.~W.; and Kavukcuoglu, K. 2016{\natexlab{a}}.
\newblock WaveNet: {A} Generative Model for Raw Audio.
\newblock In \emph{The 9th {ISCA} Speech Synthesis Workshop, (SSW)}, 125.

\bibitem[{van~den Oord et~al.(2016{\natexlab{b}})van~den Oord, Kalchbrenner, Espeholt, Kavukcuoglu, Vinyals, and Graves}]{van-den-Oord-PixelCNN}
van~den Oord, A.; Kalchbrenner, N.; Espeholt, L.; Kavukcuoglu, K.; Vinyals, O.; and Graves, A. 2016{\natexlab{b}}.
\newblock Conditional Image Generation with PixelCNN Decoders.
\newblock In \emph{Advances in Neural Information Processing Systems (NIPS)}, 4790--4798.

\bibitem[{van~den Oord, Vinyals, and Kavukcuoglu(2017)}]{Van-Den-Oord-VQ-VAE}
van~den Oord, A.; Vinyals, O.; and Kavukcuoglu, K. 2017.
\newblock Neural Discrete Representation Learning.
\newblock In \emph{Advances in Neural Information Processing Systems (NIPS)}, 6306--6315.

\bibitem[{Vaswani et~al.(2017)Vaswani, Shazeer, Parmar, Uszkoreit, Jones, Gomez, Łukasz Kaiser, and Polosukhin}]{Vaswani-Transformer}
Vaswani, A.; Shazeer, N.; Parmar, N.; Uszkoreit, J.; Jones, L.; Gomez, A.~N.; Łukasz Kaiser; and Polosukhin, I. 2017.
\newblock Attention is all you need.
\newblock In \emph{Advances in Neural Information Processing Systems (NIPS)}, 6000--6010.

\bibitem[{Wang, Yeshwanth, and Nie{\ss}ner(2021)}]{Wang-Sceneformer}
Wang, X.; Yeshwanth, C.; and Nie{\ss}ner, M. 2021.
\newblock SceneFormer: Indoor Scene Generation with Transformers.
\newblock In \emph{International Conference on 3D Vision, 3DV 2021, London, United Kingdom, December 1-3, 2021}, 106--115.

\bibitem[{Wei et~al.(2023)Wei, Ding, Park, Sajnani, Poulenard, Sridhar, and Guibas}]{Wei-LEGO-Net}
Wei, Q.~A.; Ding, S.; Park, J.~J.; Sajnani, R.; Poulenard, A.; Sridhar, S.; and Guibas, L. 2023.
\newblock LEGO-Net: Learning Regular Rearrangements of Objects in Rooms.
\newblock In \emph{Proceedings of the Computer Vision and Pattern Recognition Conference (CVPR)}, 19037--19047.

\bibitem[{Wu et~al.(2023)Wu, Wang, Gong, Liu, Xiong, Ranjan, Krishnamoorthi, Chandra, and Liu}]{Wu-PointStraightFlow}
Wu, L.; Wang, D.; Gong, C.; Liu, X.; Xiong, Y.; Ranjan, R.; Krishnamoorthi, R.; Chandra, V.; and Liu, Q. 2023.
\newblock Fast Point Cloud Generation with Straight Flows.
\newblock In \emph{Proceedings of the Computer Vision and Pattern Recognition Conference (CVPR)}, 9445--9454.

\bibitem[{Wu et~al.(2024)Wu, Feng, Wang, Xie, Dong, Miao, and Mian}]{Wu-SEK}
Wu, Z.; Feng, M.; Wang, Y.; Xie, H.; Dong, W.; Miao, B.; and Mian, A. 2024.
\newblock External Knowledge Enhanced 3D Scene Generation from Sketch.
\newblock In \emph{Computer Vision – ECCV 2024}, 286–304.

\bibitem[{Yang et~al.(2019)Yang, Huang, Hao, Liu, Belongie, and Hariharan}]{Yang-PointFlow}
Yang, G.; Huang, X.; Hao, Z.; Liu, M.; Belongie, S.~J.; and Hariharan, B. 2019.
\newblock PointFlow: 3D Point Cloud Generation With Continuous Normalizing Flows.
\newblock In \emph{Proceedings of the IEEE/CVF International Conference on Computer Vision (ICCV)}, 4540--4549.

\bibitem[{Zhai et~al.(2025)Zhai, {\"O}rnek, Chen, Liao, Di, Navab, Tombari, and Busam}]{Zhai-Echoscene}
Zhai, G.; {\"O}rnek, E.~P.; Chen, D.~Z.; Liao, R.; Di, Y.; Navab, N.; Tombari, F.; and Busam, B. 2025.
\newblock EchoScene: Indoor Scene Generation via Information Echo Over Scene Graph Diffusion.
\newblock In \emph{Computer Vision - ECCV 2024}, 167--184.

\bibitem[{Zhai et~al.(2023)Zhai, \"{O}rnek, Wu, Di, Tombari, Navab, and Busam}]{Zhai-Commonscenes}
Zhai, G.; \"{O}rnek, E.~P.; Wu, S.-C.; Di, Y.; Tombari, F.; Navab, N.; and Busam, B. 2023.
\newblock CommonScenes: generating commonsense 3D indoor scenes with scene graph diffusion.
\newblock In \emph{Advances in Neural Information Processing Systems (NeurIPS)}.

\bibitem[{Zheng and Vedaldi(2023)}]{Zheng-Online-Clustered-Codebook}
Zheng, C.; and Vedaldi, A. 2023.
\newblock Online Clustered Codebook.
\newblock In \emph{Proceedings of the IEEE/CVF International Conference on Computer Vision (ICCV)}, 22741--22750.

\end{thebibliography}

\end{document}